\documentclass[9pt,twocolumn,twoside]{revtex4-1}
\pdfoutput=1
\usepackage{graphicx}
\usepackage{amsmath}
\usepackage{textcomp}

\begin{document}

\begin{abstract}
Photonic Neural Network implementations have been gaining considerable attention as a potentially disruptive future technology.
 Demonstrating learning in large scale neural networks is essential to establish photonic machine learning substrates as viable information processing systems.
 Realizing photonic Neural Networks with numerous nonlinear nodes in a fully parallel and efficient learning hardware was lacking so far.
 We demonstrate a network of up to 2500 diffractively coupled photonic nodes, forming a large scale Recurrent Neural Network.
 Using a Digital Micro Mirror Device, we realize reinforcement learning.
 Our scheme is fully parallel, and the passive weights maximize energy efficiency and bandwidth.
 The computational output efficiently converges and we achieve very good performance.
\end{abstract}

\title{Reinforcement Learning in a large scale photonic Recurrent Neural Network}

\author{\textbf{J. Bueno$^{1}$, S. Maktoobi$^{2}$, L. Froehly$^{2}$, I. Fischer$^{1}$, M. Jacquot$^{2}$, L. Larger$^{2}$, D. Brunner$^{2,*}$}}

\affiliation{$^{1}$ Instituto de F\'{\i}sica Interdisciplinar y Sistemas Complejos, IFISC (UIB-CSIC), Campus Universitat de les Illes Balear, E-07122 Palma de Mallorca, Spain.}
\affiliation{$^{2}$ FEMTO-ST Institute/Optics Department, CNRS \& University Bourgogne Franche-Comt\'{e}, 15B avenue des Montboucons, 25030 Besançon Cedex, France. \\
	$^{*}$ Corresponding author: daniel.brunner@femto-st.fr}

\maketitle

\section{Introduction}

Multiple concepts of Neural Networks have initiated a revolution in the way we process information.
 Deep Neural Networks outperform humans in challenges previously deemed unsolvable by computers \cite{LeCun2015}.
 Among others, these systems are now capable to solve non-trivial computational problems in optics \cite{Sinha2017}.
 At the same time, Reservoir Computing (RC) emerged as a Recurrent Neural Network (RNN) concept \cite{Jaeger2004}.
 Initially, RC received substantial attention due to excellent prediction performance achieved with minimal optimization effort.
 However, quickly it was realized that RC is highly attractive for analogue hardware implementations \cite{Vandoorne2008,Appeltant2011}.

As employed by the machine learning community, Neural Networks (NNs) consist of a large number of nonlinear nodes interacting with each other.
 Evolving the NNs' state requires performing vector-matrix products with possibly millions of entries.
 Neural Network concepts therefore fundamentally benefit from parallelism.
 Consequently, photonics was identified as an attractive alternative to electronic implementation \cite{Wagner1987,Denz1998}.
 Early implementations were bulky and suffered from lack of adequate technology and NN concepts.
 This recently started to change, firstly because RC enabled a tremendous complexity-reduction of analog electronic and photonic RNNs' \cite{Appeltant2011,Duport2012,Paquot2012,Larger2012,Brunner2013a}.
 In addition, integrated photonic platforms have matured and integrated photonic Neural Networks are now feasible \cite{Shen2016}.
 Various demonstrations how a particular network of neurons can be implemented have been realized in hardware.
 Yet, Neural Networks consisting of numerous photonic nonlinear nodes combined with photonically implemented learning were so far only demonstrated in a delay systems controlled by a Field Programmable Gate Array \cite{Antonik2017}.
 Due to the time-multiplexing, delay system NNs fundamentally require such auxiliary infrastructure and computational speed suffers due to their serial nature.

While networks with multiple nodes are more challenging to implement, they offer key advantages in terms of parallelism, speed and for realizing the essential vector-matrix products.
 Here, we demonstrated a network of up to 2025 nonlinear network nodes, where each node is a pixel of a Spatial Light Modulator (SLM). 
 Recurrent and complex network connections are implemented using a Diffractive Optical Element (DOE), an intrinsically parallel and passive device \cite{Brunner2015}.
 Simulations based on the angular spectrum of plane waves show that the concept is scalable to well over 20.000 nodes.
 In a photonic RNN with $N$=900 nodes we implement learning using a digital micro-mirror device (DMD).
 The DMD is intrinsically parallel as well and, once weights have been trained, passive and energy efficient.
 Both, the coupling and learning concepts' bandwidth and power consumption is not impacted by the system's size, offering attractive scaling properties.
 Here we apply such a passive and parallel readout layer to an analogue hardware RNN, and introduce learning strategies improving performance of such systems.
 Using reinforcement learning we implement timeseries prediction with excellent performance.
 Our findings open the door to novel and versatile photonic Neural Network concepts. 

\section{Nonlinear nodes and Diffractive network}

\begin{figure}[t]
	\includegraphics[width=0.48\textwidth]{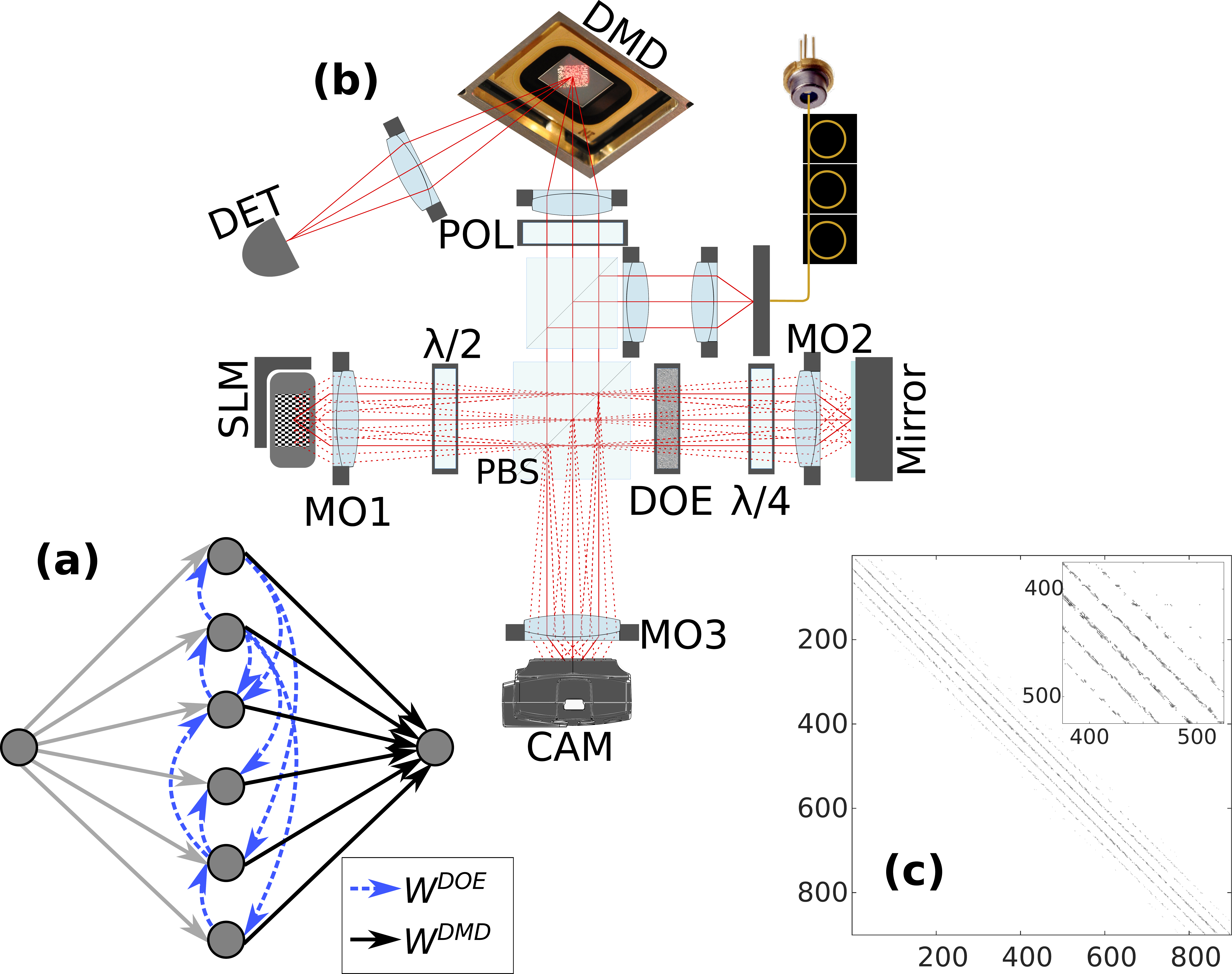}
	\caption{ \label{fig:exp_setup} 
 (a) Schematic illustration of a recurrent neural network.
 (b) Experimental setup.
 The NN state encoded on the SLM is imaged onto the camera (CAM) through a polarizing beam splitter (PBS) and a diffractive optical element (DOE).
 In this way, we realize a recurrently coupled network of Ikeda maps.
 A digital micromirror device (DMD) creates a spatially modulated image of the SLM's state.
 The integrated state, obtained via superimposing the modulated intensities, is detected, creating the photonic RNN's output.
 (c) The RNN's coupling matrix established by the DOE, with the inset showing a zoom into a smaller region.}
\end{figure} 

Figure \ref{fig:exp_setup}(a) conceptually illustrates our RNN.
 Information enters the system via a single input node, from where it is injected into a recurrently connected network of nonlinear nodes.
 The computational result is provided at the single output node after summing the network's state according to weight matrix $\mathbf{W}^{DMD}$.
 Following the RC concept, one can choose the input and recurrent internal weights randomly \cite{Jaeger2004}.
 Here, we create a complex and recurrently connected network using imaging which is spatially structured via a DOE, resulting in the internal connectivity matrix $\mathbf{W}^{DOE}$ \cite{Brunner2015}.

In Fig. \ref{fig:exp_setup}(b) we schematically illustrate our experimental setup.
 The polarization of an illumination laser (Thorlabs LP660-SF20, $\lambda$=661.2 nm, $I_{bias}=$89.69 mA, $T$=23 \textcelsius) is adjusted to s-polarization and the polarizing beam splitter cube (PBS) therefore reflects all light towards the SLM.
 By focusing the illumination laser onto the first microscope objective's (MO1, Nikon CFI Plan Achro 10X) back focal plane, the SLM (Hamamatsu X13267-01) is illuminated by a plane wave.
 The $\frac{\lambda}{2}$-plate in front of MO1 is adjusted such that the SLM operates in intensity modulation mode.
 Consequently the optical field transmitted through the PBS (p-polarization) for each SLM pixel $i$ is modulated according to
 \begin{equation}
 E_{i}=E_{i}^0\cos(2\pi \frac{x_{i}}{\kappa_{SLM}}),
 \end{equation}
 where $E_{i}^0$ and $x^{SLM}_{i}$ are the illumination and gray scale value of pixel $i$, respectively.
 $\kappa_{SLM}$ is the SLM's conversion factor between pixel gray scale and polarization rotation angle.

Ignoring for now the DOE's effect for explanatory purposes, the transmitted field is imaged (MO2, Nikon CFI Plan Achro 10X) on a mirror, and double passing through a $\frac{\lambda}{4}$-plate results in a s-polarized field.
 The PBS therefore reflects the entire optical field, which is consecutively imaged (MO2, Nikon CFI Plan Fluor 4X) on the camera (CAM, Thorlabs DCC1545M).
 We rescale the camera image via linear interpolation to fit the number of pixels of the SLM.
 This step is necessary due to (i) an imaging magnification of 2.5, and (ii) the different pixel sizes of SLM (12.5 $\mu$m) and camera (5.2 $\mu$m).
 The detected state is $x_{i}=(\alpha E_{i})^2$ with $\alpha = \frac{GS}{I^{sat}}\cdot ND$.
 Here, $I^{sat}$ is the camera's saturation intensity and $ND$ the transmission through a neutral density filter (ND) always selected such that the the dynamical range of the camera is fully exploited, while avoiding over-exposure (maximum gray scale $GS$=255).
 After multiplication with scalar $\beta$, we add a constant phase offset $\theta_{i}$ and send the resulting matrix back to the SLM.
 Ignoring the DOE's effect, each pixel therefore corresponds to an Ikeda map:
\begin{equation}
x_{i}(n+1) = GS \cos^2 ( \beta x_{i}(n) + \theta_{i}) .
\end{equation}
 We write the SLM state as vector $\mathbf{x}(n+1)$, yet in the experiment this state corresponds to a square array of SLM pixels.

Illumination wavelength, DOE (HOLOOR MS-443-650-Y-X), as well as MO1 and MO2 were chosen such that the spacing between diffractive orders matches the pixel-spacing of the SLM.
 Therefore, upon adding the DOE to the beam path, the optical field on the camera becomes $E^{C}_{i} = \sum_{j}^{N} W_{i,j}^{DOE} E_{j}$, where $\mathbf{W}^{DOE}$ is the networks coupling matrix created by the DOE.
 In Fig. \ref{fig:exp_setup}(c) we show the experimentally obtained $W^{DOE}$ for a network of 30$\times$30 nodes.
 Upon inspection of the inset one can see that locally connectivity strengths vary significantly.
 This is due to each pixel illuminating a DOE area comparable to the DOE's lowest spatial frequency.
 As this area shifts slightly from pixel to pixel, the intensity distribution between diffractive orders varies.
 This intended effect inherently creates the heterogeneous photonic network topology needed for computation \cite{Jaeger2004}.
 Finally, the photonic RNN's state $\mathbf{x}(n+1)$ is given by
\begin{multline}
x_{i}(n+1) = GS \cdot \cos^{2}( \beta \cdot \alpha\left( \sum_{j}^{N} W_{i,j}^{DOE} E_{j}\right)^2 \\ + \gamma W_i^{inj} u(n+1) + \theta_{i}) .
\end{multline} 
 Here $u(n+1)$ is the information to be injected into the RNN and $\gamma$ is the signal injection strength.
 Matlab is used to control all instruments, update the network state and to inject the input information weighted by matrix $\gamma W_i^{inj}$.
 The  overall update rate of the entire system is $\sim$5 Hz.
 Currently, the maximum size of networks we can realize consist of $\sim$2500 nodes, which is limited by the imaging setup's field of view and not by the concept itself.

\section{Network readout weights}

\begin{figure}[t]
	\includegraphics[width=0.48\textwidth]{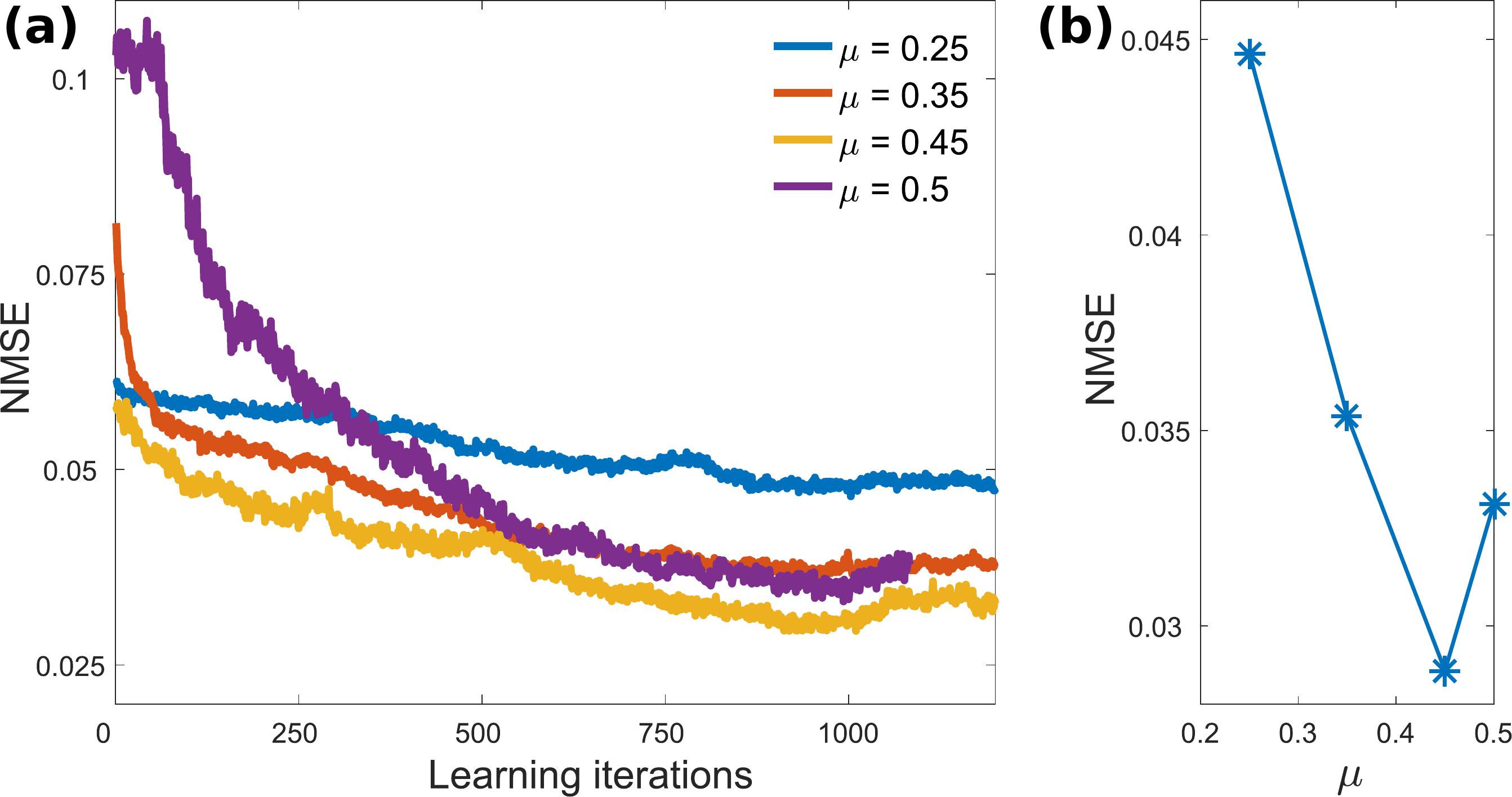}
	\caption{ \label{fig:theta_division} (a) Learning curves for a photonic RNN with two phase offsets $\phi_i$ across the network.
	A fraction of $(1-\mu)$ nodes have a response function with a negative slope ($\Theta_0=0.17\pi$), the remaining nodes experience a response function with a positive slope ($\Theta_0=0.43\pi$).
	(b) An almost symmetric distribution of phase offsets, resulting in positive and negative node responses, benefits computational performance.}
\end{figure}

Having created a recurrent photonic neural network driven by external data, the final to information processing is to adjust the system such that it performs the required computation.
 This is typically achieved by modifying connection weights according to some learning routine.
 Inspired by the RC concept, we constrain learning induced weight adjustment to the readout layer.
 Our 900 RNN nodes are spatially distributed, and we therefore can use a simple lens (Thorlabs AC254-400-B) to image the SLM's state onto a commercial array of micro mirrors (DLi4120 XGA, pitch 13.68 $\mu$m).
 Micro mirrors can be flipped between $\pm$12 \textdegree, such that for -12 \textdegree the optical signal is directed to a detector.
 The detector´s photo current then corresponds to the RNN output.
 With $\mathbf{W}^{DMD}$ as the readout weight vector, the RNN output becomes
\begin{equation}\label{eq:readout}
y^{out}(n+1) = \delta \left( \mathbf{W}^{DMD} (1 - \mathbf{x}(n+1)) \right).
\end{equation}
\noindent Here, $\delta$ relates the power recorded by the power meter (Thorlabs S150C and PM100A) to the SLM state.
 The signal directed towards the DMD is orthogonally polarized compared to the one directed to the camera, resulting in $\mathbf{x}^{DMD}(n) = 1 - \mathbf{x}(n)$.
 In the experiment weight vector $\mathbf{W}^{DMD}$ corresponds to a square matrix, which can be seen in Fig. \ref{fig:exp_setup}(b).
 The image labeled DMD shows a typical configuration of the DMD with $\mathbf{W}^{DMD}$.
 As the contribution of each node is either on or off, $\mathbf{W}^{DMD}$ consists of Boolean entries only.
 Weights are not temporal modulations as in delay system implementations of RC \cite{Antonik2017}, and therefore can be implemented by passive attenuations in reflection or transmission.
 Such passive processes are energy efficient and typically do not results in a bandwidth limitation.
 In this specific implementation, once trained, mirrors can simply remain in their position, and if mechanically fixed, would not consume additional energy.
 Also, readout eq. \ref{eq:readout} is optically performed for all elements in parallel.

\section{Photonic learning}

It is now the task of a learning algorithm to tailor $\mathbf{W}^{DMD}$ such that signal $y^{out}(n+1)$ approximates a target value as good as possible.
 In our experiment, we employ a version of reinforcement learning.
 The learning input signal is injected after inverting the weight assigned to one node.
 The error $\varepsilon_{k}$ of signal $y^{out}_{k}(n+1)$ obtained for configuration $\mathbf{W}^{DMD}_{k}$ is then compare to the error $\varepsilon_{k-1}$, where $k$ is the index of learning iterations.
 If the error is reduced, we keep DMD configuration $\mathbf{W}^{DMD}_{k}$, if not, we revert back to $\mathbf{W}^{DMD}_{k-1}$ and invert a different weight.
 The weight to be updated is determined by the largest entry´s $W^{select,max}_{k}$ position $l_{k}$ according to
\begin{eqnarray}
\mathbf{W}^{select}_{k} =& rand(N)\cdot \mathbf{W}^{bias}, \\
\left[ l_{k}, W^{select,max}_{k} \right] =& max(\mathbf{W}^{select}_{k}), \\
W^{DMD}_{k, l_{k}} =& \neg  (W^{DMD}_{k-1,l_{k}}), \\
\mathbf{W}^{bias} = \frac{1}{N} + \mathbf{W}^{bias},& W^{bias}_{l_{k}} = 0. 
\end{eqnarray} 
$ rand(N)$ creates a random vector with $N$ entries, and at the start $\mathbf{W}^{DMD}_{1}$ and $\mathbf{W}^{bias}$ are randomly initialized.
 $\mathbf{W}^{bias}$ acts as a bias vector, whose values are increased by $\frac{1}{N}$ each learning iteration while the bias belonging to the most recently updated weight is set to zero.
 This results in a randomized selection process with a bias away from inverting recently updated weights.
 In simulations we found that reinforcement learning including such a bias showed significantly faster learning convergence.
 As a task to be performed by our system, we chose nonlinear time series prediction.
 The injected signal $u(n+1)$ is the chaotic Mackey-Glass (MG) sequence \cite{Jaeger2004}, and the RNN's learning target is $y^{T}(n+1) = u(n+2)$, the one-time-step-prediction of the MG system.
 Parameters of the temporal MG sequence where identical to \cite{Bueno2016}, using an integration step size of 0.1.
 For determining the error $\varepsilon_{k}$ we discarded the first 30 data points due to their transient nature.
 The RNN´s remaining output sequence was then inverted, its offset subtracted and normalized by its standard deviation, creating signal $\tilde{y}^{out}_{k}$.
 The error was measured by $\varepsilon_{k} = \sigma(y^{T} - \tilde{y}^{out}_{k})$, where $\sigma$ is the standard deviation and $\varepsilon_{k}$ therefore corresponds to the normalized mean square error (NMSE).

At this stage we would like to stress a significant difference between neural networks emulated on digital electronic computers and our photonic hardware implementation.
 In our system, all connection weights are positive, and $\mathbf{W}^{DMD}$ is boolean.
 This restricts the functional space available for approximating the targeted input-output transformation.
 As a result, first evaluations of the learning procedure and prediction of the MG series suffered from minor performance.
 However, we were able to mitigate this restriction by harnessing the non-monotonous slope of the $\cos^2$ nonlinearity.
 We randomly divided the offest phases $\theta_{i} |_{i=1...N}$, resulting in nodes with negative and positive slope of their response function.
 We chose $\Theta_0 = 42 \hat{=} 0.17\pi$ and $\Theta_0+\Delta\Theta = 106\hat{=} 0.43\pi$, respectively, with a probability of $1-\mu$ for $\theta_{i} = \Theta_0$.
 As RNN-states and $\mathbf{W}^{DOE}$ entries are exclusively positive, the nonlinear transformation of nodes with $\theta_{i} = \Theta_0$ is predominantly along a positive slope, for $\theta_{i} = \Theta_0 + \Delta\Theta$ along a negative slope.
 This enables the reinforcement learning procedure to select from nonlinear transformations with positive and negative slopes.
 We used feedback $\beta = 0.8$ and injection gain $\gamma = 0.4$, and learning curves for various ratios ($\mu = \left[ 0.25, 0.35, 0.45, 0.5 \right]$) are shown in Fig. \ref{fig:theta_division} (a).
 They reveal a strong impact of this symmetry breaking.
 Optimum performance for each $\mu$ is shown in \ref{fig:theta_division}(b).
 Best performance is found for an RNN operating around almost equally distributed operating points at $\mu = 0.45$.
 This demonstrates that the absence of negative weights in $\mathbf{W}^{DMD}$, $\mathbf{W}^{DOE}$ and $\mathbf{x}$ can be partially compensated for by incorporating nonlinear transformations with positive as well as negative slopes.
 This result is of high significance for optical neural networks, which, e.g. motivated by robustness considerations, renounce making use of the optical phase to implement negative weights.

\begin{figure}[t]
	\includegraphics[width=0.5\textwidth]{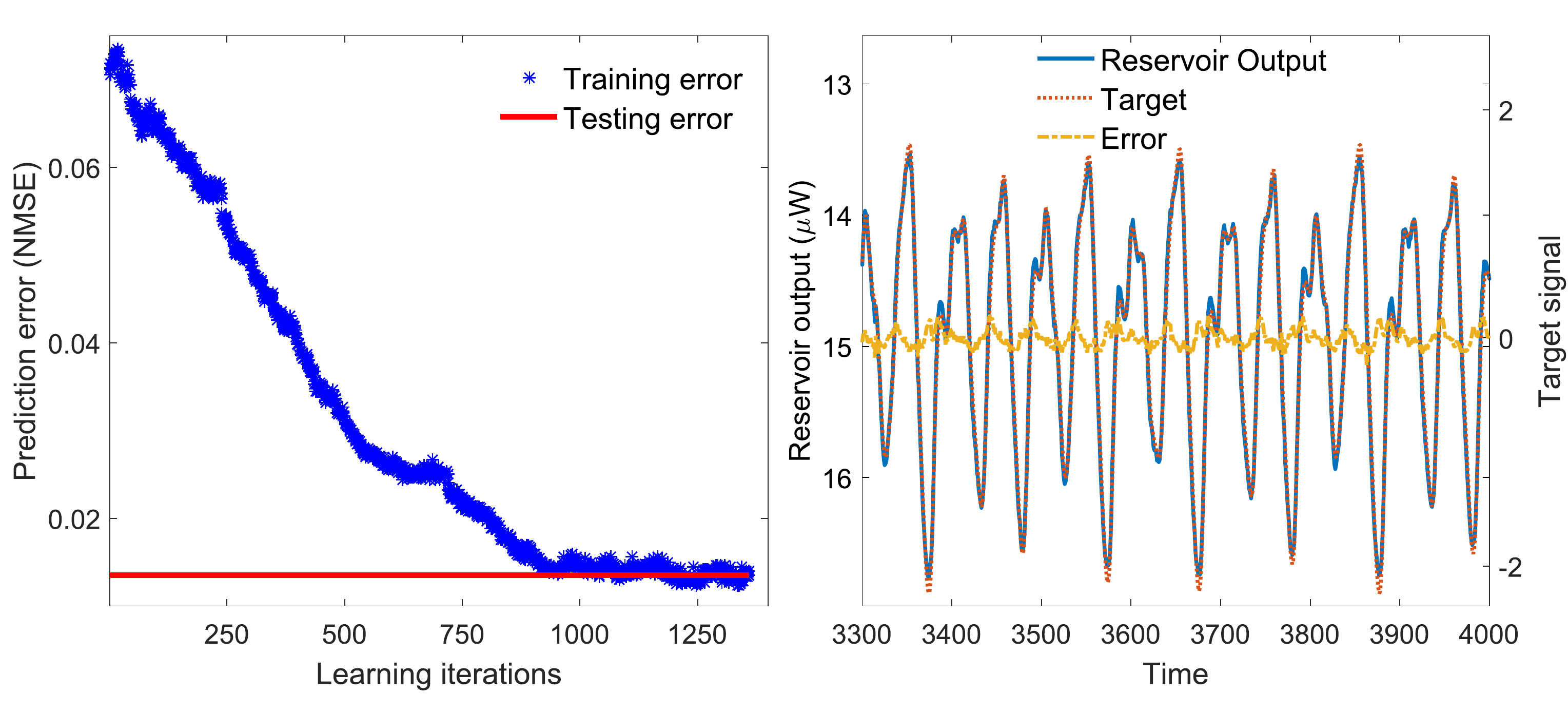}
	\caption{ \label{fig:traintest} (a) Learning performance at optimal parameters ($\beta$=0.8, $\gamma$=0.4, $\mu$=0.45).
	(b) The photonic RNN's predicted output in $\mu$W (blue line) can hardly be differentiated from the prediction target (red dots).
	Prediction error $\varepsilon$ is given by the yellow dashed data.}
\end{figure}

We further optimized our system's performance by scanning the remaining parameters $\beta$ and $\gamma$.
 In Fig. \ref{fig:traintest} (a) we show the error convergence under optimized global conditions for a training sample size of 500 steps (blue stars).
 The error efficiently reduces, and finally stabilizes at $\varepsilon \approx 0.013$.
 Considering learning is limited to Boolean readout weights this is an excellent result.
 After training, the prediction performance is evaluated further on a sequence of 4500 datapoints consecutive data points which were not part of the training dataset.
 As indicated by the red line in the same panel, the testing error matches the training error.
 We can therefore conclude that our photonic RNN successfully generalized the underlying target system's properties.
 The excellent prediction performance can be appreciated in Fig. \ref{fig:traintest} (b).
 Data belonging to the left y-axis (blue line) shows the recorded output power, while on the right y-axis (red dots) we show the normalized prediction target signal.
 A difference between both is hardly visible, and the prediction error $\varepsilon$ (yellow dashed line) is small.
 Down-sampling the injected signals by 3 creates condition identical to \cite{Bueno2016,Ortin2015}.
 Under these conditions our error ($\varepsilon=0.042$) is larger by a factor of 2.2 relative to a delay RC based on a semiconductor laser \cite{Bueno2016} and by 6.5 relative to a Mach-Zehnder modulator based setup \cite{Ortin2015}.
 These comparisons have to be evaluated in the light of the significantly increased level of hardware implementation in our current setup.
 In \cite{Bueno2016,Ortin2015}, readout weights were applied digitally in an off-line procedure using weights with double precision.
 In \cite{Ortin2015} a strong impact of digitization resolution on the computational performance was identified, suggesting that $\varepsilon$ can be significantly reduced by increasing the resolution of $\mathbf{W}^{DMD}$.
 
\section{Conclusion}

We demonstrated a photonic RNN consisting of hundreds of photonic nonlinear nodes and the implementation of photonic reinforcement learning.
 Using a simple Boolean valued readout implemented with a DMD, we trained our system to predict the chaotic MG sequence.
 The resulting prediction error is very low despite of the Boolean readout weights.

In our work we demonstrate how symmetry breaking inside the RNN can compensate for exclusively positive intensities in our analogue neural networks systems.
 These results resolve a complication of general importance to neural networks implemented in analog hardware.
 Hardware-implemented networks and readout weights based on physical devices open the door to a new class of experiments, i.e. evaluating the robustness and efficiency of learning strategies in fully implemented analogue neural networks.
 The final step, a photonic realization of the input, should be straight forward, as it only requires a complex spatial distribution of the input information.
 Finally, our system is not limited to the reported slow opto-electronic system.
 Extremely fast all-optical systems can be realized employing the same concept since we intentionally implemented a $4f$ architecture to allow for self-coupling \cite{Brunner2015}.

\section*{Funding Information}

The authors acknowledge the support of the Region Bourgogne Franche-Comt\'{e}.
 This work was supported by Labex ACTION program (contract ANR-11-LABX-0001-0) and via the Volkswagen Foundation NeuroQNet project.

\section*{Acknowledgments}

The authors would like to thank Christian Markus Dietrich for valuable contributions to earlier versions of the setup.

\bibliography{bibfile}

\begin{thebibliography}{16}%
\makeatletter
\providecommand \@ifxundefined [1]{%
 \@ifx{#1\undefined}
}%
\providecommand \@ifnum [1]{%
 \ifnum #1\expandafter \@firstoftwo
 \else \expandafter \@secondoftwo
 \fi
}%
\providecommand \@ifx [1]{%
 \ifx #1\expandafter \@firstoftwo
 \else \expandafter \@secondoftwo
 \fi
}%
\providecommand \natexlab [1]{#1}%
\providecommand \enquote  [1]{``#1''}%
\providecommand \bibnamefont  [1]{#1}%
\providecommand \bibfnamefont [1]{#1}%
\providecommand \citenamefont [1]{#1}%
\providecommand \href@noop [0]{\@secondoftwo}%
\providecommand \href [0]{\begingroup \@sanitize@url \@href}%
\providecommand \@href[1]{\@@startlink{#1}\@@href}%
\providecommand \@@href[1]{\endgroup#1\@@endlink}%
\providecommand \@sanitize@url [0]{\catcode `\\12\catcode `\$12\catcode
  `\&12\catcode `\#12\catcode `\^12\catcode `\_12\catcode `\%12\relax}%
\providecommand \@@startlink[1]{}%
\providecommand \@@endlink[0]{}%
\providecommand \url  [0]{\begingroup\@sanitize@url \@url }%
\providecommand \@url [1]{\endgroup\@href {#1}{\urlprefix }}%
\providecommand \urlprefix  [0]{URL }%
\providecommand \Eprint [0]{\href }%
\providecommand \doibase [0]{http://dx.doi.org/}%
\providecommand \selectlanguage [0]{\@gobble}%
\providecommand \bibinfo  [0]{\@secondoftwo}%
\providecommand \bibfield  [0]{\@secondoftwo}%
\providecommand \translation [1]{[#1]}%
\providecommand \BibitemOpen [0]{}%
\providecommand \bibitemStop [0]{}%
\providecommand \bibitemNoStop [0]{.\EOS\space}%
\providecommand \EOS [0]{\spacefactor3000\relax}%
\providecommand \BibitemShut  [1]{\csname bibitem#1\endcsname}%
\let\auto@bib@innerbib\@empty
\bibitem [{\citenamefont {LeCun}\ \emph {et~al.}(2015)\citenamefont {LeCun},
  \citenamefont {Bengio},\ and\ \citenamefont {Hinton}}]{LeCun2015}%
  \BibitemOpen
  \bibfield  {author} {\bibinfo {author} {\bibfnamefont {Y.}~\bibnamefont
  {LeCun}}, \bibinfo {author} {\bibfnamefont {Y.}~\bibnamefont {Bengio}}, \
  and\ \bibinfo {author} {\bibfnamefont {G.}~\bibnamefont {Hinton}},\ }\href
  {\doibase 10.1038/nature14539} {\bibfield  {journal} {\bibinfo  {journal}
  {Nature}\ }\textbf {\bibinfo {volume} {521}},\ \bibinfo {pages} {436}
  (\bibinfo {year} {2015})}\BibitemShut {NoStop}%
\bibitem [{\citenamefont {Sinha}\ \emph {et~al.}(2017)\citenamefont {Sinha},
  \citenamefont {Lee}, \citenamefont {Li},\ and\ \citenamefont
  {Barbastathis}}]{Sinha2017}%
  \BibitemOpen
  \bibfield  {author} {\bibinfo {author} {\bibfnamefont {A.}~\bibnamefont
  {Sinha}}, \bibinfo {author} {\bibfnamefont {J.}~\bibnamefont {Lee}}, \bibinfo
  {author} {\bibfnamefont {S.}~\bibnamefont {Li}}, \ and\ \bibinfo {author}
  {\bibfnamefont {G.}~\bibnamefont {Barbastathis}},\ }\href {\doibase
  10.1364/OPTICA.4.001117} {\bibfield  {journal} {\bibinfo  {journal} {Optica}\
  }\textbf {\bibinfo {volume} {4}},\ \bibinfo {pages} {1117} (\bibinfo {year}
  {2017})}\BibitemShut {NoStop}%
\bibitem [{\citenamefont {Jaeger}\ and\ \citenamefont
  {Haas}(2004)}]{Jaeger2004}%
  \BibitemOpen
  \bibfield  {author} {\bibinfo {author} {\bibfnamefont {H.}~\bibnamefont
  {Jaeger}}\ and\ \bibinfo {author} {\bibfnamefont {H.}~\bibnamefont {Haas}},\
  }\href {\doibase 10.1126/science.1091277} {\bibfield  {journal} {\bibinfo
  {journal} {Science (New York, N.Y.)}\ }\textbf {\bibinfo {volume} {304}},\
  \bibinfo {pages} {78} (\bibinfo {year} {2004})}\BibitemShut {NoStop}%
\bibitem [{\citenamefont {Vandoorne}\ \emph {et~al.}(2008)\citenamefont
  {Vandoorne}, \citenamefont {Dierckx}, \citenamefont {Schrauwen},
  \citenamefont {Verstraeten}, \citenamefont {Baets}, \citenamefont
  {Bienstman},\ and\ \citenamefont {{Van Campenhout}}}]{Vandoorne2008}%
  \BibitemOpen
  \bibfield  {author} {\bibinfo {author} {\bibfnamefont {K.}~\bibnamefont
  {Vandoorne}}, \bibinfo {author} {\bibfnamefont {W.}~\bibnamefont {Dierckx}},
  \bibinfo {author} {\bibfnamefont {B.}~\bibnamefont {Schrauwen}}, \bibinfo
  {author} {\bibfnamefont {D.}~\bibnamefont {Verstraeten}}, \bibinfo {author}
  {\bibfnamefont {R.}~\bibnamefont {Baets}}, \bibinfo {author} {\bibfnamefont
  {P.}~\bibnamefont {Bienstman}}, \ and\ \bibinfo {author} {\bibfnamefont
  {J.}~\bibnamefont {{Van Campenhout}}},\ }\href {\doibase
  10.1364/OE.16.011182} {\bibfield  {journal} {\bibinfo  {journal} {Optics
  Express}\ }\textbf {\bibinfo {volume} {16}},\ \bibinfo {pages} {11182}
  (\bibinfo {year} {2008})}\BibitemShut {NoStop}%
\bibitem [{\citenamefont {Appeltant}\ \emph {et~al.}(2011)\citenamefont
  {Appeltant}, \citenamefont {Soriano}, \citenamefont {{Van der Sande}},
  \citenamefont {Danckaert}, \citenamefont {Massar}, \citenamefont {Dambre},
  \citenamefont {Schrauwen}, \citenamefont {Mirasso},\ and\ \citenamefont
  {Fischer}}]{Appeltant2011}%
  \BibitemOpen
  \bibfield  {author} {\bibinfo {author} {\bibfnamefont {L.}~\bibnamefont
  {Appeltant}}, \bibinfo {author} {\bibfnamefont {M.~C.}\ \bibnamefont
  {Soriano}}, \bibinfo {author} {\bibfnamefont {G.}~\bibnamefont {{Van der
  Sande}}}, \bibinfo {author} {\bibfnamefont {J.}~\bibnamefont {Danckaert}},
  \bibinfo {author} {\bibfnamefont {S.}~\bibnamefont {Massar}}, \bibinfo
  {author} {\bibfnamefont {J.}~\bibnamefont {Dambre}}, \bibinfo {author}
  {\bibfnamefont {B.}~\bibnamefont {Schrauwen}}, \bibinfo {author}
  {\bibfnamefont {C.~R.}\ \bibnamefont {Mirasso}}, \ and\ \bibinfo {author}
  {\bibfnamefont {I.}~\bibnamefont {Fischer}},\ }\href {\doibase
  10.1038/ncomms1476} {\bibfield  {journal} {\bibinfo  {journal} {Nature
  Communications}\ }\textbf {\bibinfo {volume} {2}},\ \bibinfo {pages} {468}
  (\bibinfo {year} {2011})}\BibitemShut {NoStop}%
\bibitem [{\citenamefont {Wagner}\ and\ \citenamefont
  {Psaltis}(1987)}]{Wagner1987}%
  \BibitemOpen
  \bibfield  {author} {\bibinfo {author} {\bibfnamefont {K.}~\bibnamefont
  {Wagner}}\ and\ \bibinfo {author} {\bibfnamefont {D.}~\bibnamefont
  {Psaltis}},\ }\href {\doibase 10.1364/AO.26.005061} {\bibfield  {journal}
  {\bibinfo  {journal} {Applied Optics}\ }\textbf {\bibinfo {volume} {26}},\
  \bibinfo {pages} {5061} (\bibinfo {year} {1987})}\BibitemShut {NoStop}%
\bibitem [{\citenamefont {Denz}(1998)}]{Denz1998}%
  \BibitemOpen
  \bibfield  {author} {\bibinfo {author} {\bibfnamefont {C.}~\bibnamefont
  {Denz}},\ }\href {\doibase 10.1007/978-3-663-12272-2} {\emph {\bibinfo
  {title} {{Optical Neural Networks}}}}\ (\bibinfo  {publisher} {Springer
  Vieweg},\ \bibinfo {address} {Wiesbaden},\ \bibinfo {year}
  {1998})\BibitemShut {NoStop}%
\bibitem [{\citenamefont {Duport}\ \emph {et~al.}(2012)\citenamefont {Duport},
  \citenamefont {Schneider}, \citenamefont {Smerieri}, \citenamefont
  {Haelterman},\ and\ \citenamefont {Massar}}]{Duport2012}%
  \BibitemOpen
  \bibfield  {author} {\bibinfo {author} {\bibfnamefont {F.}~\bibnamefont
  {Duport}}, \bibinfo {author} {\bibfnamefont {B.}~\bibnamefont {Schneider}},
  \bibinfo {author} {\bibfnamefont {A.}~\bibnamefont {Smerieri}}, \bibinfo
  {author} {\bibfnamefont {M.}~\bibnamefont {Haelterman}}, \ and\ \bibinfo
  {author} {\bibfnamefont {S.}~\bibnamefont {Massar}},\ }\href
  {http://www.ncbi.nlm.nih.gov/pubmed/23037429} {\bibfield  {journal} {\bibinfo
   {journal} {Optics Express}\ }\textbf {\bibinfo {volume} {20}},\ \bibinfo
  {pages} {22783} (\bibinfo {year} {2012})}\BibitemShut {NoStop}%
\bibitem [{\citenamefont {Paquot}\ \emph {et~al.}(2012)\citenamefont {Paquot},
  \citenamefont {Duport}, \citenamefont {Smerieri}, \citenamefont {Dambre},
  \citenamefont {Schrauwen}, \citenamefont {Haelterman},\ and\ \citenamefont
  {Massar}}]{Paquot2012}%
  \BibitemOpen
  \bibfield  {author} {\bibinfo {author} {\bibfnamefont {Y.}~\bibnamefont
  {Paquot}}, \bibinfo {author} {\bibfnamefont {F.}~\bibnamefont {Duport}},
  \bibinfo {author} {\bibfnamefont {A.}~\bibnamefont {Smerieri}}, \bibinfo
  {author} {\bibfnamefont {J.}~\bibnamefont {Dambre}}, \bibinfo {author}
  {\bibfnamefont {B.}~\bibnamefont {Schrauwen}}, \bibinfo {author}
  {\bibfnamefont {M.}~\bibnamefont {Haelterman}}, \ and\ \bibinfo {author}
  {\bibfnamefont {S.}~\bibnamefont {Massar}},\ }\href {\doibase
  10.1038/srep00287} {\bibfield  {journal} {\bibinfo  {journal} {Scientific
  Reports}\ }\textbf {\bibinfo {volume} {2}},\ \bibinfo {pages} {287} (\bibinfo
  {year} {2012})}\BibitemShut {NoStop}%
\bibitem [{\citenamefont {Larger}\ \emph {et~al.}(2012)\citenamefont {Larger},
  \citenamefont {Soriano}, \citenamefont {Brunner}, \citenamefont {Appeltant},
  \citenamefont {Gutierrez}, \citenamefont {Pesquera}, \citenamefont
  {Mirasso},\ and\ \citenamefont {Fischer}}]{Larger2012}%
  \BibitemOpen
  \bibfield  {author} {\bibinfo {author} {\bibfnamefont {L.}~\bibnamefont
  {Larger}}, \bibinfo {author} {\bibfnamefont {M.~C.}\ \bibnamefont {Soriano}},
  \bibinfo {author} {\bibfnamefont {D.}~\bibnamefont {Brunner}}, \bibinfo
  {author} {\bibfnamefont {L.}~\bibnamefont {Appeltant}}, \bibinfo {author}
  {\bibfnamefont {J.~M.}\ \bibnamefont {Gutierrez}}, \bibinfo {author}
  {\bibfnamefont {L.}~\bibnamefont {Pesquera}}, \bibinfo {author}
  {\bibfnamefont {C.~R.}\ \bibnamefont {Mirasso}}, \ and\ \bibinfo {author}
  {\bibfnamefont {I.}~\bibnamefont {Fischer}},\ }\href {\doibase
  10.1364/OE.20.003241} {\bibfield  {journal} {\bibinfo  {journal} {Optics
  Express}\ }\textbf {\bibinfo {volume} {20}},\ \bibinfo {pages} {3241}
  (\bibinfo {year} {2012})}\BibitemShut {NoStop}%
\bibitem [{\citenamefont {Brunner}\ \emph {et~al.}(2013)\citenamefont
  {Brunner}, \citenamefont {Soriano}, \citenamefont {Mirasso},\ and\
  \citenamefont {Fischer}}]{Brunner2013a}%
  \BibitemOpen
  \bibfield  {author} {\bibinfo {author} {\bibfnamefont {D.}~\bibnamefont
  {Brunner}}, \bibinfo {author} {\bibfnamefont {M.~C.}\ \bibnamefont
  {Soriano}}, \bibinfo {author} {\bibfnamefont {C.~R.}\ \bibnamefont
  {Mirasso}}, \ and\ \bibinfo {author} {\bibfnamefont {I.}~\bibnamefont
  {Fischer}},\ }\href {\doibase 10.1038/ncomms2368} {\bibfield  {journal}
  {\bibinfo  {journal} {Nature Communications}\ }\textbf {\bibinfo {volume}
  {4}},\ \bibinfo {pages} {1364} (\bibinfo {year} {2013})}\BibitemShut
  {NoStop}%
\bibitem [{\citenamefont {Shen}\ \emph {et~al.}(2017)\citenamefont {Shen},
  \citenamefont {Harris}, \citenamefont {Skirlo}, \citenamefont {Prabhu},
  \citenamefont {Baehr-Jones}, \citenamefont {Hochberg}, \citenamefont {Sun},
  \citenamefont {Zhao}, \citenamefont {Larochelle}, \citenamefont {Englund},\
  and\ \citenamefont {Soljacic}}]{Shen2016}%
  \BibitemOpen
  \bibfield  {author} {\bibinfo {author} {\bibfnamefont {Y.}~\bibnamefont
  {Shen}}, \bibinfo {author} {\bibfnamefont {N.~C.}\ \bibnamefont {Harris}},
  \bibinfo {author} {\bibfnamefont {S.}~\bibnamefont {Skirlo}}, \bibinfo
  {author} {\bibfnamefont {M.}~\bibnamefont {Prabhu}}, \bibinfo {author}
  {\bibfnamefont {T.}~\bibnamefont {Baehr-Jones}}, \bibinfo {author}
  {\bibfnamefont {M.}~\bibnamefont {Hochberg}}, \bibinfo {author}
  {\bibfnamefont {X.}~\bibnamefont {Sun}}, \bibinfo {author} {\bibfnamefont
  {S.}~\bibnamefont {Zhao}}, \bibinfo {author} {\bibfnamefont {H.}~\bibnamefont
  {Larochelle}}, \bibinfo {author} {\bibfnamefont {D.}~\bibnamefont {Englund}},
  \ and\ \bibinfo {author} {\bibfnamefont {M.}~\bibnamefont {Soljacic}},\
  }\href {\doibase 10.1038/nphoton.2017.93} {\bibfield  {journal} {\bibinfo
  {journal} {Nature Photonics}\ }\textbf {\bibinfo {volume} {11}},\ \bibinfo
  {pages} {441} (\bibinfo {year} {2017})},\ \Eprint
  {http://arxiv.org/abs/1610.02365} {arXiv:1610.02365} \BibitemShut {NoStop}%
\bibitem [{\citenamefont {Antonik}\ \emph {et~al.}(2017)\citenamefont
  {Antonik}, \citenamefont {Haelterman},\ and\ \citenamefont
  {Massar}}]{Antonik2017}%
  \BibitemOpen
  \bibfield  {author} {\bibinfo {author} {\bibfnamefont {P.}~\bibnamefont
  {Antonik}}, \bibinfo {author} {\bibfnamefont {M.}~\bibnamefont {Haelterman}},
  \ and\ \bibinfo {author} {\bibfnamefont {S.}~\bibnamefont {Massar}},\ }\href
  {\doibase 10.1007/s12559-017-9459-3} {\bibfield  {journal} {\bibinfo
  {journal} {Cognitive Computation}\ ,\ \bibinfo {pages} {1}} (\bibinfo {year}
  {2017})}\BibitemShut {NoStop}%
\bibitem [{\citenamefont {Brunner}\ and\ \citenamefont
  {Fischer}(2015)}]{Brunner2015}%
  \BibitemOpen
  \bibfield  {author} {\bibinfo {author} {\bibfnamefont {D.}~\bibnamefont
  {Brunner}}\ and\ \bibinfo {author} {\bibfnamefont {I.}~\bibnamefont
  {Fischer}},\ }\href@noop {} {\bibfield  {journal} {\bibinfo  {journal}
  {Optics Letters}\ }\textbf {\bibinfo {volume} {40}},\ \bibinfo {pages} {3854}
  (\bibinfo {year} {2015})}\BibitemShut {NoStop}%
\bibitem [{\citenamefont {Bueno}\ \emph {et~al.}(2016)\citenamefont {Bueno},
  \citenamefont {Brunner}, \citenamefont {Soriano},\ and\ \citenamefont
  {Fischer}}]{Bueno2016}%
  \BibitemOpen
  \bibfield  {author} {\bibinfo {author} {\bibfnamefont {J.}~\bibnamefont
  {Bueno}}, \bibinfo {author} {\bibfnamefont {D.}~\bibnamefont {Brunner}},
  \bibinfo {author} {\bibfnamefont {M.}~\bibnamefont {Soriano}}, \ and\
  \bibinfo {author} {\bibfnamefont {I.}~\bibnamefont {Fischer}},\ }\href
  {\doibase 10.1364/OE.25.002401} {\bibfield  {journal} {\bibinfo  {journal}
  {Optics Express}\ }\textbf {\bibinfo {volume} {25}},\ \bibinfo {pages} {2401}
  (\bibinfo {year} {2016})}\BibitemShut {NoStop}%
\bibitem [{\citenamefont {Ort{\'{i}}n}\ \emph {et~al.}(2015)\citenamefont
  {Ort{\'{i}}n}, \citenamefont {Pesquera}, \citenamefont {Brunner},
  \citenamefont {San-Mart{\'{i}}n}, \citenamefont {Fischer}, \citenamefont
  {Mirasso},\ and\ \citenamefont {Guti{\'{e}}rrez}}]{Ortin2015}%
  \BibitemOpen
  \bibfield  {author} {\bibinfo {author} {\bibfnamefont {M.~C.}\ \bibnamefont
  {Ort{\'{i}}n}, \bibfnamefont {S.and~Soriano}}, \bibinfo {author}
  {\bibfnamefont {L.}~\bibnamefont {Pesquera}}, \bibinfo {author}
  {\bibfnamefont {D.}~\bibnamefont {Brunner}}, \bibinfo {author} {\bibfnamefont
  {D.}~\bibnamefont {San-Mart{\'{i}}n}}, \bibinfo {author} {\bibfnamefont
  {I.}~\bibnamefont {Fischer}}, \bibinfo {author} {\bibfnamefont {C.~R.}\
  \bibnamefont {Mirasso}}, \ and\ \bibinfo {author} {\bibfnamefont {J.~M.}\
  \bibnamefont {Guti{\'{e}}rrez}},\ }\href {\doibase 10.1038/srep14945}
  {\bibfield  {journal} {\bibinfo  {journal} {Scientific Reports}\ }\textbf
  {\bibinfo {volume} {5}},\ \bibinfo {pages} {14945} (\bibinfo {year}
  {2015})}\BibitemShut {NoStop}%
\end{thebibliography}%

\end{document}